# Gray-Level Image Transitions Driven by Tsallis Entropic Index

## Amelia Carolina Sparavigna

Department of Applied Science and Technology, Politecnico di Torino, Italy.

**Abstract:** The maximum entropy principle is largely used in thresholding and segmentation of images. Among the several formulations of this principle, the most effectively applied is that based on Tsallis non-extensive entropy. Here, we discuss the role of its entropic index in determining the thresholds. When this index is spanning the interval (0,1), for some images, the values of thresholds can have large leaps. In this manner, we observe abrupt transitions in the appearance of corresponding bi-level or multi-level images. These gray-level image transitions are analogous to order or texture transitions observed in physical systems, transitions which are driven by the temperature or by other physical quantities.

## 1. Introduction

In 1988, a new formulation of entropy was postulated, in the framework of a non-extensive thermodynamics, by Constantino Tsallis [1]. Like the scaling functions of the universal multifractals that inspired it [1-3], this entropy depends on a dimensionless parameter, the entropic index, usually indicated by $q$. In the limit $q \to 1$, it is recovering the expression of Boltzmann-Gibbs-Shannon entropy.

By a simple analysis of the scientific literature, it is easy to see that, today, the Tsallis entropy is becoming the corner stone of several processing methods, in particular when applied to medical and fuzzy processing [4]. This happens because the Tsallis entropy is used for those image segmentations which are based on the maximum entropy principle [5]. Let us remember that segmentation is the processing task which separates the pixels in those pertaining to objects and in those pertaining to the background, helping then in extracting some of the image features [6].

In a recent paper [4], we have discussed the role of Tsallis entropy in bi-level and multi-level image thresholding. In this processing, a gray-scale image is converted in a bi-level or multi-level image. Here we will discuss the role of the entropic index, the parameter $q$, in giving the values of thresholds. We will see that, changing the value of index $q$, we can have an interesting behaviour of the corresponding threshold values.

In fact, when $q$ is spanning continuously the interval (0,1), in some cases, the value of a threshold can have a large jump. As a consequence, we observe an abrupt transition in the appearance of the resulting bi-level or multi-level thresholding. Therefore, we have an "image transition" between different gray bi-level or multi-level images. This is analogous to what we can observe in order or texture transitions displayed by physical systems. In this case, the systems are governed by their related free energies, and the transitions are driven by the temperature or by other physical quantities.

## 2. Tsallis entropy and image segmentation

In [7], a survey of thresholding is given, which is discussing and categorizing methods into some groups, distinguished by means of the information the algorithms are manipulating. Among them, we find methods based on histograms and maximum entropy principle [8]. In [9], the authors used a method similar to the maximum entropy sum method [8], but applying Tsallis entropy. To the best of our knowledge, they were the first researches to use this entropy in image processing.

Let us consider an image having $k$ gray levels. An image of size $X \times Y$ is a matrix of the form $[f(x,y) \mid x = 1, 2, \ldots, X; \quad y = 1, 2, \ldots, Y]$, where $f(x,y)$ is the gray value of pixel located at point $(x,y)$. The set of all gray values $\{0, 1, 2, \ldots, k\}$ usually has $k = 255$. We need a distribution of probabilities, $p_1, p_2, \ldots, p_k$ for the gray levels: for it, we can use the normalized histogram $h_i = n_i / n_{tot}$; $n_i$ is the number of pixels with gray value $i$ and $n_{tot}$ is the number of all pixels in the image (in Figure 1, we show three examples of histograms of images). The Tsallis entropy of the image is:

$$S_q^I = \frac{1}{q-1}\left\{1 - \sum_{i=1}^{k} p_i^q\right\}$$
$$= \frac{1}{q-1}\left\{1 - \sum_{i=1}^{k} h_i^q\right\} \quad (1)$$

The entropy is a function of entropic index $q$.

## 3. Bi-level thresholding

Let us assume a bi-level threshold $t$ for the gray levels. In [9], two classes had been introduced, $A$ and $B$, and their probability distributions:

$$A: \frac{p_1}{p_A}, \frac{p_2}{p_A}, \ldots, \frac{p_t}{p_A} \quad (2)$$

$$B: \frac{p_{t+1}}{p_B}, \frac{p_{t+2}}{p_B}, \ldots, \frac{p_k}{p_B} \quad (3)$$

$$p_A = \sum_{i=1}^{t} p_i; \; p_B = \sum_{i=t+1}^{k} p_i \quad (4)$$

The Tsallis entropies, one for each distribution, are given by:

$$S_q^A(t) = \frac{1}{q-1}\left\{1 - \sum_{i=1}^{t}\left(\frac{p_i}{p_A}\right)^q\right\} \quad (5)$$

$$S_q^B(t) = \frac{1}{q-1}\left\{1 - \sum_{i=t+1}^{k}\left(\frac{p_i}{p_B}\right)^q\right\} \quad (6)$$

The total Tsallis entropy is:

$$S_q(t) = S_q^A(t) + S_q^B(t) + (1-q)S_q^A(t)S_q^B(t) \quad (7)$$

When this entropy, which is a function of threshold $t$, is maximized, the corresponding gray level $t$ is considered the optimum threshold value [9].

In the gray bi-level thresholding, we have a resulting processed image, which is a black and white image. The output image is created as in the following: if pixels have a gray tone larger than the threshold, they become white. If pixels have a lower value, they become black.

However, its appearance is depending on the given threshold. And this threshold is depending on the entropic index $q$. Two examples of thresholding are given in Figure 2 and 3, on two images (stem cells and blood cells) of Fig.1. In the case of Figure 2, we can see that the value of the threshold is increasing, more or less, continuously without large leaps. In the case of the Figure 3, the value of the threshold does not change when $q$ is spanning the interval (0,1). Let us note that in a bi-level image, the red cells are not visible, unless a value of q quite close to zero is used.

## 4. An image transition

In the Figures 2 and 3, we can see the Tsallis entropy as a function of the threshold $t$. Where this function has a maximum, the corresponding value of $t$ is considered the best for thresholding. Since the final result depends on $q$, this index can be used as an adjustable parameter and can play an important role as a tuning parameter [9]. This is quite important in some cases. Let us consider, for instance, the Figure 4, where we can see the original image, which is the lower image of Figure 1, and the effects of its thresholding.

In the lower part of the Figure 4, we can see the behavior of the threshold as a function of $q$. This function has an abrupt jump at a critical value $q$=0.515. To this jump, it corresponds an analogous abrupt change in the appearance of the corresponding bi-level image. Moreover, as a consequence of this change, we have that the two bi-level images are providing different information. One is able to distinguish the blackest cells.

We are defining this behavior of gray bi-level images as an "image transition", because it is analogous, for instance, to the texture transitions in liquid crystals [10,11]. In these transitions, the texture of a liquid crystal cell, observed in polarized liquid microscopy, changes abruptly its appearance, because the material, driven by the temperature, had changed its free energy. Then, in images, the Tsallis entropy is playing the role of the free energy and its entropic index the role of temperature.

## 5. Three-level thresholding
In the multi-level thresholding, some classes are introduced, separated by $t_1, t_2, \ldots, t_m$ thresholds. The probability distributions of them and the related entropies have been discussed in [4].
In a three-level thresholding, we are considering two thresholds $t_1, t_2$, and we have three classes:

$$(1): \frac{p_1}{P_1}, \frac{p_2}{P_1}, \ldots, \frac{p_{t_1}}{P_1} \qquad (8)$$

$$(2): \frac{p_{t_1+1}}{P_2}, \frac{p_{t_1+2}}{P_2}, \ldots, \frac{p_{t_2}}{P_2} \qquad (9)$$

$$(3): \frac{p_{t_2+1}}{P_3}, \frac{p_{t_2+2}}{P_3}, \ldots, \frac{p_k}{P_3} \qquad (10)$$

In (8), (9) and (10), we have:

$$P_1 = \sum_{i=1}^{t_1} p_i ; \ P_2 = \sum_{i=t_1+1}^{t_2} p_i ; \ P_3 = \sum_{i=t_2+1}^{k} p_i \qquad (11)$$

Tsallis entropies are:

$$S_q^{(1)} = \frac{1}{q-1} \left\{ 1 - \sum_{i=1}^{t_1} \left( \frac{p_i}{P_1} \right)^q \right\} \qquad (12)$$

$$S_q^{(2)} = \frac{1}{q-1} \left\{ 1 - \sum_{i=t_1+1}^{t_2} \left( \frac{p_i}{P_2} \right)^q \right\} \qquad (13)$$

$$S_q^{(3)} = \frac{1}{q-1} \left\{ 1 - \sum_{i=t_2+1}^{k} \left( \frac{p_i}{P_3} \right)^q \right\} \qquad (14)$$

The total Tsallis entropy is:

$$
\begin{aligned}
S_q = {} & S_q^{(1)} + S_q^{(2)} + S_q^{(3)} \\
& + (1-q) S_q^{(1)} S_q^{(2)} + (1-q) S_q^{(1)} S_q^{(3)} \\
& + (1-q) S_q^{(2)} S_q^{(3)} + (1-q)^2 S_q^{(1)} S_q^{(2)} S_q^{(3)}
\end{aligned} \qquad (15)
$$

In [12], the authors proposed this multi-level thresholding method based on Tsallis entropy for image segmentation. They used an interesting artificial bee colony approach to reduce the time of processing. In [13] and [14], a multi-level thresholding with Tsallis entropy had been addressed too.
In the Figures 5 and 6, there are three examples of three-level thresholding. For each image, we optimized two thresholds. In the Figures we can see the behavior of entropy too: in red (dark gray), we have the entropy as a function of threshold $t_1$ for different values of the other threshold; in green (light gray), we have the same for threshold $t_2$. Let us stress that the values of the two thresholds are a function of the entropic index.
In the case of the image with stem cells (upper panel of Figure 5), we do not observe, when $q$ spans interval $(0,1)$, any jump in the values of the thresholds. But in the case of the other two images (Figure 5, lower part, and Figure 6), we observe again an image transition. In the Figure 5, we have that, for the image of blood cells, a critical value of $q$, about 0.665, exists. The visibility of the red cells disappears above this critical value of $q$.

## 6. Conclusion
In this paper, we have discussed the role of Tsallis entropic index in determining the bi-level and three-level thresholding. For some images, the values of thresholds can have a jump, when the entropic index is spanning interval $(0,1)$. This is provoking an abrupt transition in the appearance of the corresponding output images. We can define this behavior as an "image transition". Of course, the investigation of image transitions can be further extended to the general multi-level thresholding. The gray-level image transitions are analogous to order or texture transitions observed in physical systems.

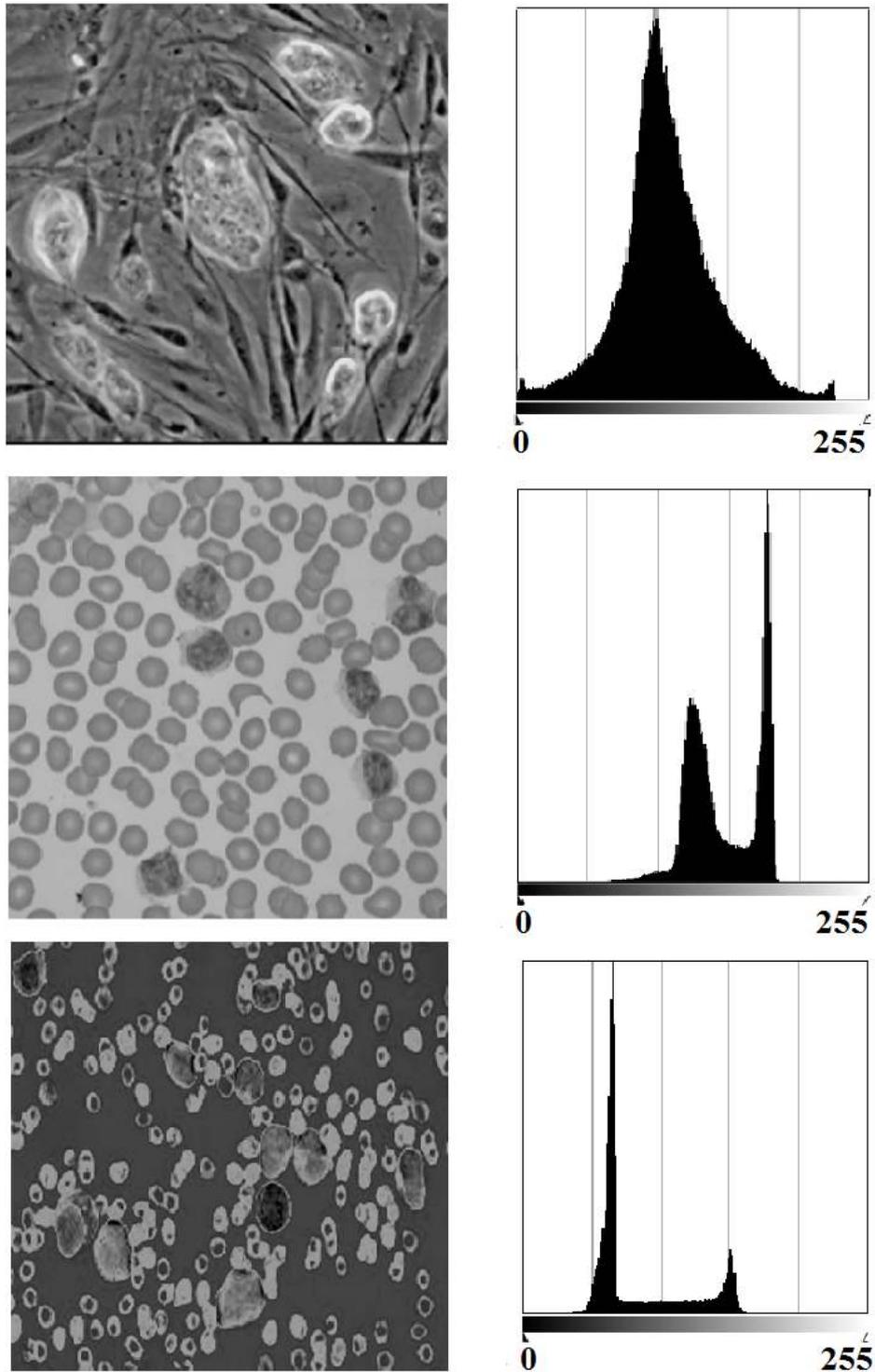

Figure 1 - The images show mouse embryonic stem cells (up) and blood cells (middle) (Courtesy Wikimedia Commons) in gray tones. The lower panel shows an image of blood cells processed with GIMP (the GNU Image Manipulation Program), to change its gray tones. On the right, we can see the histograms of the corresponding images, with tones ranging from 0 to 255. The histogram allows obtaining the probability of a gray tone and, from it, the Tsallis entropy of the image.

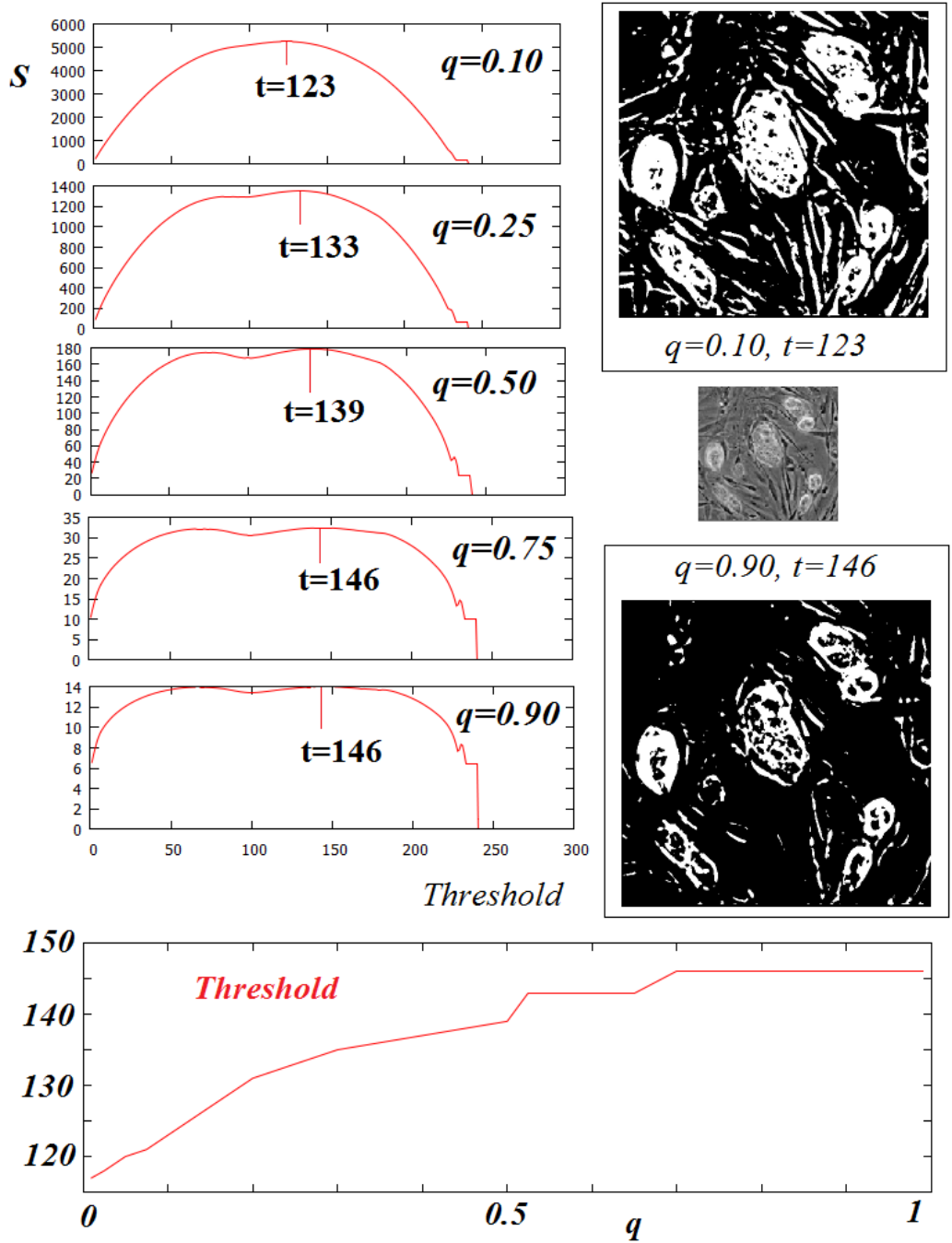

Figure 2 - Effects of a bi-level thresholding obtained by maximizing Tsallis entropy, on the image of stem cells. Pixels having a gray tone larger than the threshold become white; pixels having a lower value become black. Tsallis entropy $S$ as a function of threshold is also given for different values of the entropic index $q$. In each panel, the maximum value is marked. In the lower part of the Figure, we can see the optimal threshold as a function of $q$ in the interval (0,1).

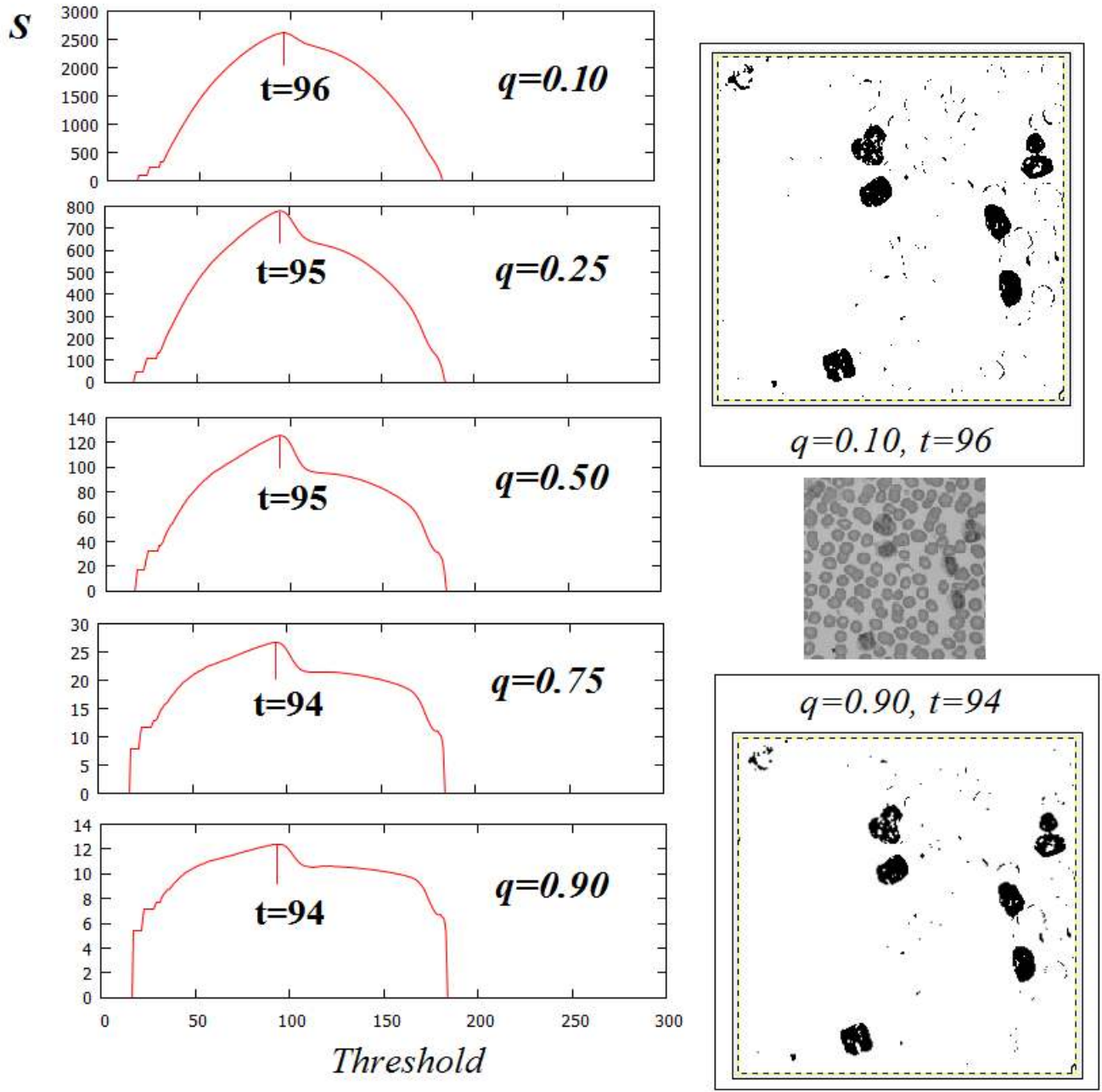

Figure 3 - Effects of a bi-level thresholding obtained by maximizing Tsallis entropy, on the image of blood cells. Pixels having a gray tone larger than the threshold become white; pixels having a lower value become black. Tsallis entropy $S$ as a function of threshold is also given for different values of the entropic index $q$. Note that in these bi-level images, the red cells are not visible. When $q$ is spanning interval (0,1), the value of the optimized threshold is quite constant, for q>0.01 . Tsallis entropy is not able to show in the bi-level image the red cells of blood, unless a value of $q$ quite close to zero is used.

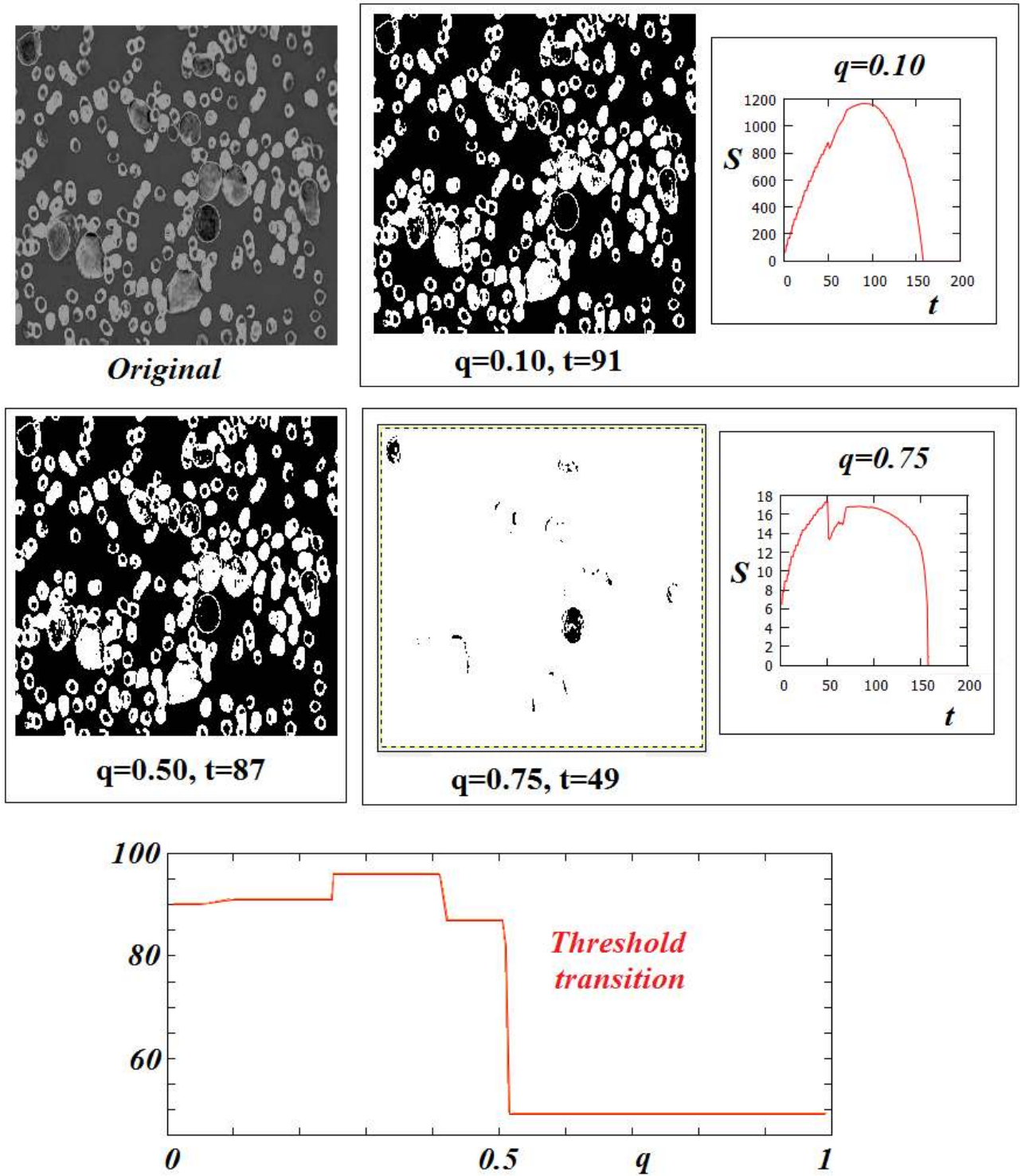

Figure 4 - Bi-level thresholding obtained  by maximizing Tsallis entropy, in the case of the lower image of Figure 1. Tsallis entropy $S$ as a function of threshold is also given for different values of the entropic index $q$. In the lower part of the Figure, we can see  the optimized threshold as a function of $q$ in the interval (0,1). Note that we have a gray-level image transition at $q$=0.505. The bi-level image, which is almost black when $q$ is below this critical value, becomes almost white when $q$ is above  it and the red cells disappear.

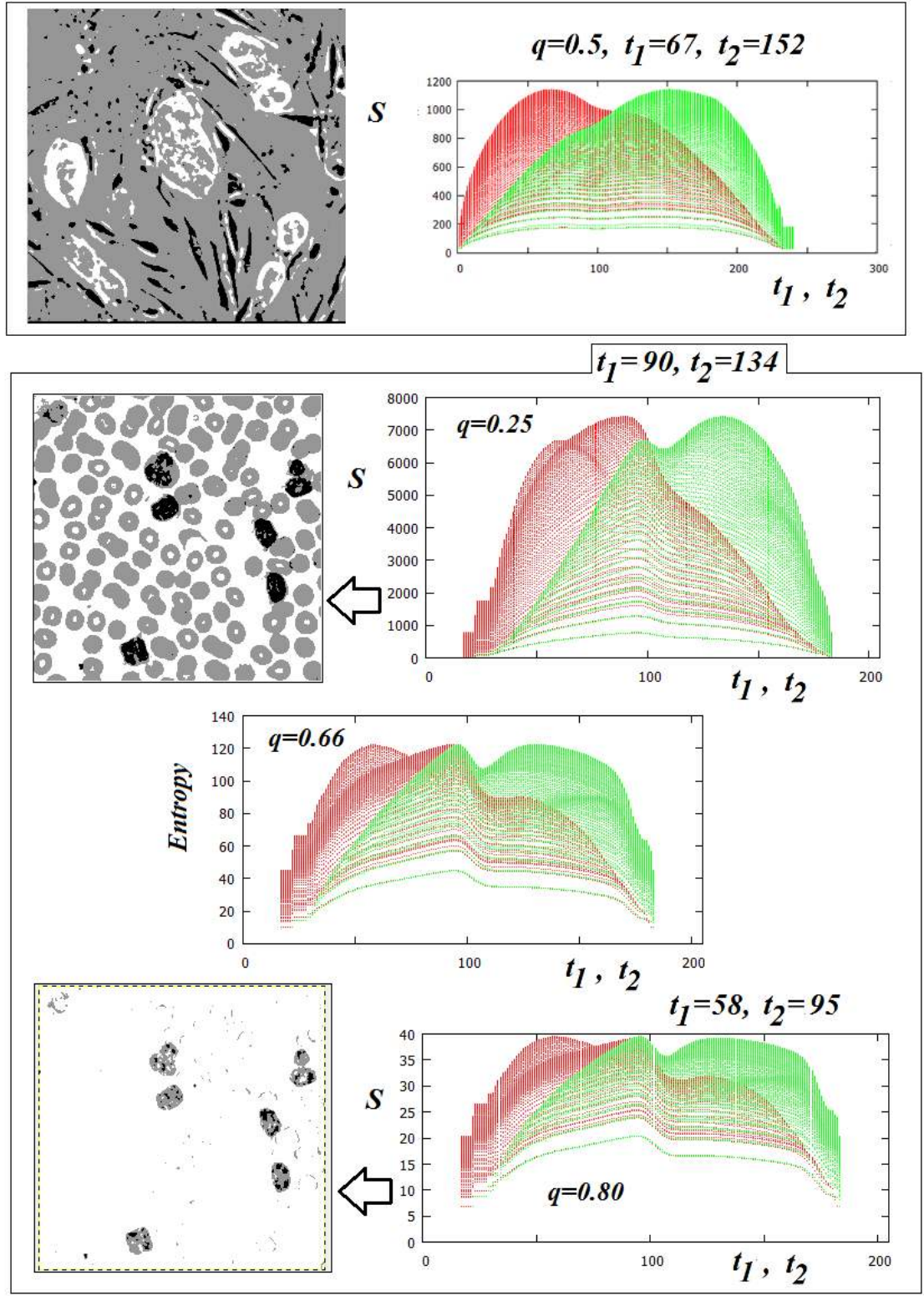

Figure 5 - Examples of three-level thresholding using Tsallis entropy on the images of stem cells and blood cells. We can see also the behavior of the entropy: in red (dark gray) we have the entropy as a function of threshold $t_1$, for different values of the other threshold; in green (light gray), we show the same for threshold $t_2$. Note that a gray-level image transition exists for the image of blood cells. The critical value of $q$ is about 0.665. The visibility of the red cells disappears above the critical value.

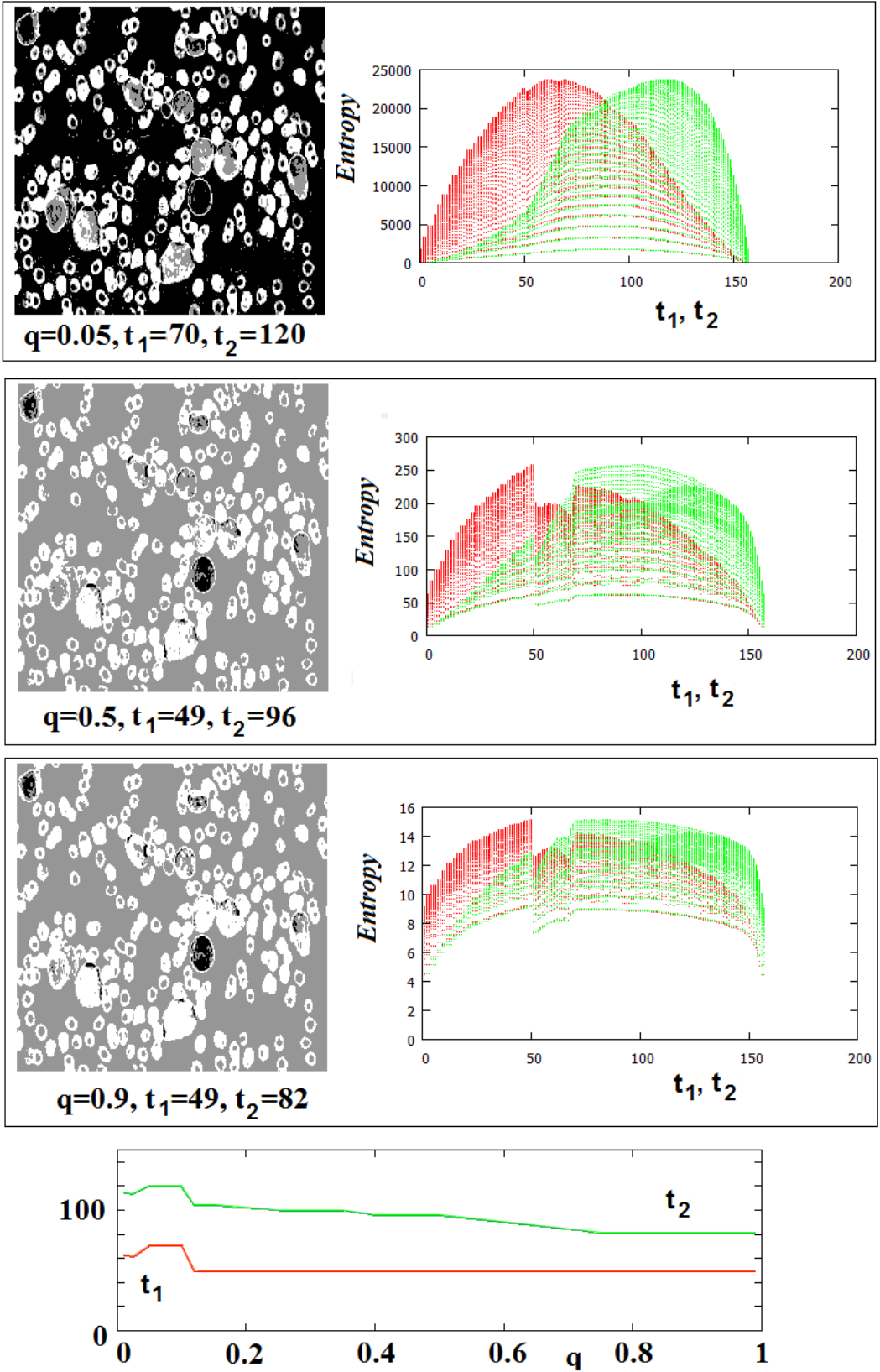

Figure 6 - Three-level thresholding using Tsallis entropy on the lower image of blood cells of Figure 1. We can see the behavior of entropy: in red (dark gray) we have the entropy as a function of threshold $t_1$, for different values of the other threshold; in green (light gray), we show the same for threshold $t_2$. Let us note again an image transition. The critical value of $q$ is 0.10, as we can see from the plot of optimized thresholds as functions of the entropic index.